\documentclass[11pt,a4paper]{article}
\usepackage[hyperref]{naaclhlt2019}
\usepackage{times}
\usepackage{latexsym}

\usepackage{url}

\usepackage{amsmath}
\usepackage{subcaption}
\usepackage{graphicx}
\usepackage{hyperref}

\aclfinalcopy 

\title{Revisiting Visual Grounding}

\author{Erik Conser \\
  Computer Science Department \\
  Portland State University\\
  {\tt econser@pdx.edu} \\\And
  Kennedy Hahn \\ 
  Computer Science Department \\ 
  Portland State University\\
  {\tt kehahn@pdx.edu} \\
 \AND
  Chandler M. Watson \\
  Computer Science Department \\ 
   Stanford University \\
  {\tt chandler.watson@stanford.edu} \\\And
  Melanie Mitchell \\
  Computer Science Department \\ 
  Portland State University\\
  and Santa Fe Institute \\
  {\tt mm@pdx.edu} \\}

\date{}

\begin{document}
\maketitle

\begin{abstract}
We revisit a particular visual grounding method: the ``Image Retrieval
Using Scene Graphs'' (IRSG) system of Johnson et
al.\ \shortcite{Johnson2015}.  Our experiments indicate that the
system does not effectively use its learned object-relationship
models.  We also look closely at the IRSG dataset, as well as the
widely used Visual Relationship Dataset (VRD) that is adapted from it.  We find
that these datasets exhibit biases that allow methods that ignore
relationships to perform relatively well.  We also describe several
other problems with the IRSG dataset, and report on experiments using
a subset of the dataset in which the biases and other problems are
removed.  Our studies contribute to a more general effort: that of
better understanding what machine learning methods that combine
language and vision actually learn and what popular datasets
actually test.

\end{abstract}

\section{Introduction}

\begin{figure*}[h]
\center
\includegraphics[width=6in]{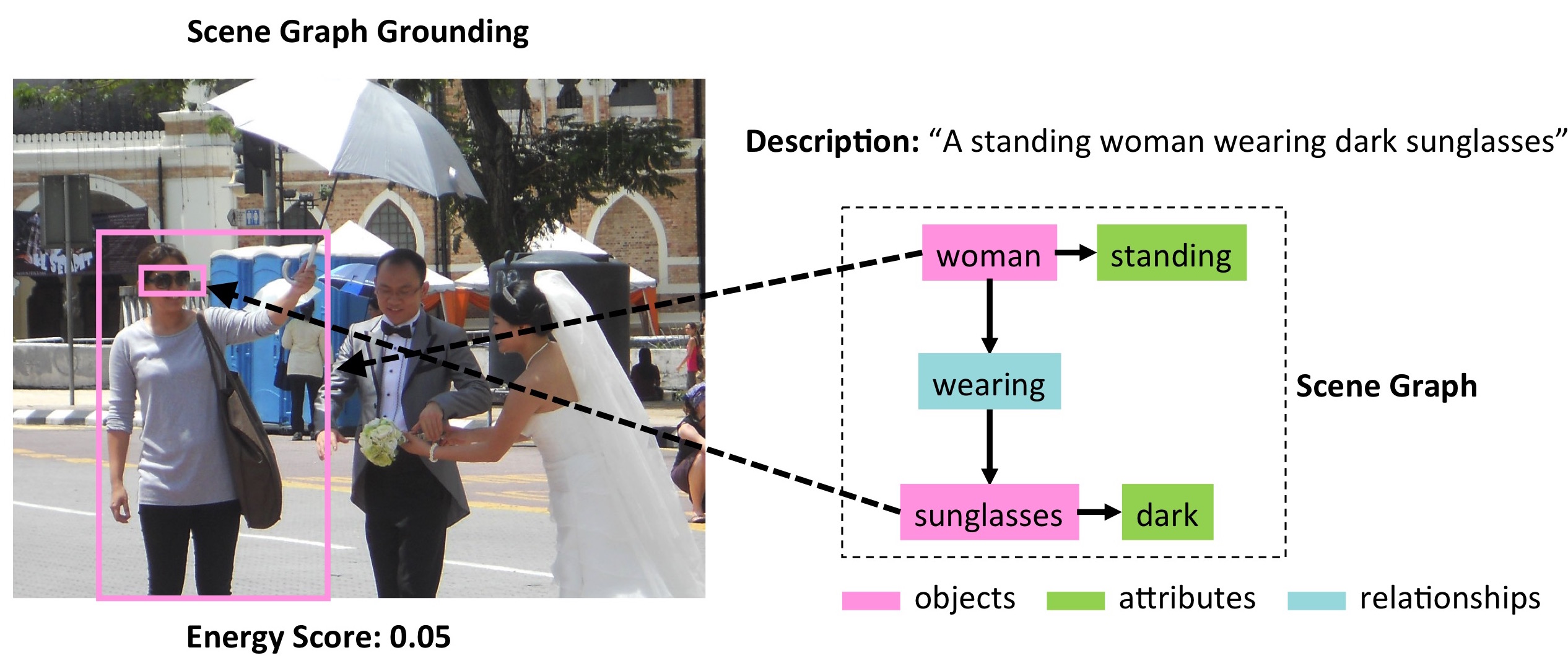}
\caption{An example of the scene-graph-grounding task of Johnson et al.\ 
  \shortcite{Johnson2015}.  Right: A phrase represented as a scene graph.
  Left: A candidate grounding of the scene graph in a test image, here
  yielding a low energy score (lower is better).}
\label{SGTask}
\end{figure*}

\begin{figure*}[h]
\center
\includegraphics[width=6in]{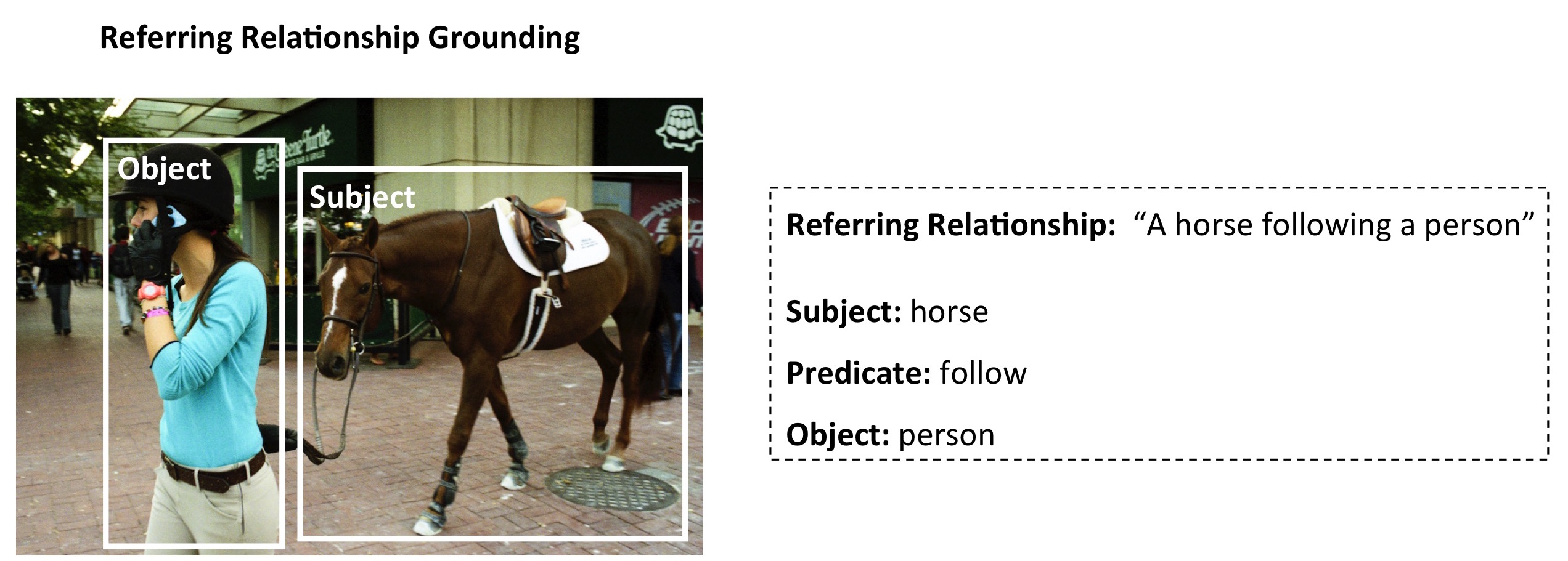}
\caption{An example of the referring-relationship-grounding task of
  Krishna et al.\ \shortcite{Krishna2018}.  Right: A phrase broken into subject, predicate,
  and object categories.  Left: a candidate grounding of subject and
  object in a test image.}
\label{RRTask}
\end{figure*}

{\it Visual grounding} is the general task of locating the components
of a structured description in an image.  In the visual-grounding
literature, the structured description is often a natural-language
phrase that has been parsed as a {\it scene graph} or as a {\it
  subject-predicate-object} triple.  As one example of a
visual-grounding challenge, Figure~\ref{SGTask} illustrates the
``Image Retrieval using Scene Graphs'' (IRSG) task 
\cite{Johnson2015}.  Here the sentence ``A standing woman wearing dark
sunglasses'' is converted to a scene-graph representation (right) with
nodes corresponding to objects, attributes, and relationships.  Given
a scene graph and an input image, the grounding task is to create
bounding boxes corresponding to the specified objects, such that the
located objects have the specified attributes and relationships
(left).  A final energy score reflects the quality of the match
between the scene graph and the located boxes (lower is better), and can be used to rank
images in a retrieval task.  A second example of visual grounding, illustrated in Figure~\ref{RRTask}, is
the ``Referring Relationships'' (RR) task of Krishna et al.\ \shortcite{Krishna2018}.
Here, a sentence (e.g., ``A horse following a person'') is represented
as a subject-predicate-object triple (``horse'', ``following'',
``person''). Given a triple and an input image, the task is to
create bounding boxes corresponding to the named subject and object,
such that the located boxes fit the specified predicate.  Visual
grounding tasks---at the intersection of vision and language---have
become a popular area of research in machine learning, with the
potential of improving automated image editing, captioning, retrieval, 
and question-answering, among other tasks.

While deep neural networks have produced impressive progress in object
detection, visual-grounding tasks remain highly challenging. On the
language side, accurately transforming a natural language phrase to a
structured description can be difficult.  On the vision side, the
challenge is to learn---in a way that can be generalized---visual
features of objects and attributes as well as flexible models of
spatial and other relationships, and then to apply these models to
figure out which of a given object class (e.g., {\it woman}) is the one
referred to, sometimes locating small objects and recognizing
hard-to-see attributes (e.g., {\it dark} vs. {\it clear} glasses).  To
date, the performance of machine learning systems on visual-grounding
tasks with real-world datasets has been relatively low compared to
human performance.

In addition, some in the machine-vision community have questioned the
effectiveness of popular datasets that have been developed to evaluate the
performance of systems on visual grounding tasks like the ones
illustrated in Figures~\ref{SGTask} and~\ref{RRTask}.  Recently Cirik
et al.\ \shortcite{Cirik2018b} showed that for the widely used dataset Google-Ref \cite{Mao2016}, 
the task of grounding referring expressions has exploitable biases:
for example, a system that predicts only object categories---ignoring
relationships and attributes---still performs well on this task. Jabri
et al.\ \shortcite{Jabri2016} report related biases in visual
question-answering datasets.  \\ \indent In this paper we re-examine
the visual grounding approach of Johnson et
al.\ \shortcite{Johnson2015} to determine how well this system is
actually performing scene-graph grounding.  In particular, we compare
this system with a simple baseline method to test if the original
system is using information from object relationships, as claimed by
Johnson et al.\ \shortcite{Johnson2015}. In addition, we investigate
possible biases and other problems with the dataset used by Johnson et
al.\ \shortcite{Johnson2015}, a version of which has also been used in
many later studies.  We briefly survey related work in visual
grounding, and discuss possible future studies in this area.

\section{Image Retrieval Using Scene Graphs}

\subsection{Methods}
The ``Image Retrieval Using Scene Graphs'' (IRSG) method 
\cite{Johnson2015} performs the task illustrated in
Figure~\ref{SGTask}: given an input image and a scene graph, output a
{\it grounding} of the scene graph in the image and an accompanying
energy score.  The grounding consists of a set of bounding boxes, 
each one corresponding to an object named in the scene graph, with the
goal that the grounding gives the the best possible fit to the
objects, attributes, and relationships specified in the scene graph.  Note
that the system described in \cite{Johnson2015} does not perform any
linguistic analysis; it assumes that a natural-language description
has already been transformed into a scene graph.

The IRSG system is trained on a set of
human-annotated images in which bounding boxes are labeled with object
categories and attributes, and pairs of bounding boxes are labeled
with relationships.  The system learns appearance models for all
object and attribute categories in the training set, and relationship
models for all training-set relationships.  The appearance model for
object categories is learned as a convolutional neural network (CNN), which
inputs an bounding box from an image and outputs a probability
distribution over all object categories.  The appearance
model for object attributes is also learned as a CNN; it inputs an image bounding
box and outputs a probability distribution over all attribute
categories.  The pairwise spatial relationship models are learned as
Gaussian mixture models (GMMs); each GMM inputs a pair of bounding
boxes from an image and outputs a probability density reflecting how
well the GMM judges the input boxes to fit the model's corresponding
spatial relationship (e.g., ``woman wearing sunglasses''). Details of the
training procedures are given in \cite{Johnson2015}.

After training is completed, the IRSG system can be run on test
images. Given a test image and a scene graph, IRSG attempts to ground
the scene graph in the image as follows.  First the system creates a
set of candidate bounding boxes using the Geodesic Object Proposal
method \cite{Krahenbuhl2014}.  The object and attribute CNNs are then
used to assign probability distributions over all object and attribute categories
to each candidate bounding box.  Next, for each relationship in the
scene graph, the GMM corresponding to that relationship assigns a
probability density to each pair of candidate bounding boxes.  The
probability density is calibrated by Platt scaling \cite{Platt2000} to
provide a value representing the probability that the given pair of
boxes is in the specified relationship.

Finally, these object and relationship probabilities are used to
configure a conditional random field, implemented as factor
graph. The objects and attributes are unary factors in the factor
graph, each with one value for each image bounding box.  The relationships
are binary factors, with one value for each pair of bounding boxes.
This factor graph represents the probability distribution of
groundings conditioned on the scene graph and bounding boxes.  Belief
propagation \cite{Andres2012} is then run on the factor graph to determine which 
candidate bounding boxes produce the lowest-energy grounding of the
given scene graph.  The output of the system is this grounding, along
with its energy.  The lower the energy, the better the predicted fit
between the image and the scene graph.

To use the IRSG system in image retrieval, with a query represented as
a scene graph, the IRSG system applies the grounding procedure for the
given scene graph to every image in the test set, and ranks the
resulting images in order of increasing energy. The highest ranking
(lowest energy) images can be returned as the results of the query.

Johnson et al.\ \shortcite{Johnson2015} trained and tested the IRSG method
on an image dataset consisting of 5,000 images, split into 4,000
training images and 1,000 testing images.  The objects, attributes,
and relationships in each image were annotated by Amazon Mechanical
Turk workers; the authors created scene graphs that captured the
annotations.  IRSG was tested on two types of scene-graph queries:
full and partial.  Each full scene-graph query was a highly detailed
description of a single image in the test set---the average full scene
graph consisted of 14 objects, 19 attributes, and 22 relationships.
The partial scene graphs were generated by examination of subgraphs of
the full scene graphs.  Each combination of two objects, one relation,
and one or two attributes was drawn from each full scene graph, and any
partial scene graph that was found at least five times was added to the
collection of partial queries.  Johnson et al.\ 
randomly selected 119 partial queries to constitute the test set for
partial queries.   

\subsection{Original Results}   
Johnson et al.\ \shortcite{Johnson2015} used a ``recall at $k$'' metric to
measure their their system's image retrieval performance. In experiments on both
full and partial scene-graph queries, the authors found that their
method outperformed several baselines. In particular, it
outperformed---by a small degree---two ``ablated'' forms of their
method: the first in which only object probabilities were used
(attribute and relationship probabilities were ignored), and the
second in which both object and attribute probabilities were used but 
relationship probabilities were ignored.

\section{Revisiting IRSG}

We obtained the IRSG code from the authors \cite{Johnson2015}, and
attempted to replicate their reported results on the partial scene
graphs.  (Our study included only the partial scene graphs, which
seemed to us to be a more realistic use case for image retrieval than the
complex full graphs, each of which described only one image in the
set.)  We performed additional experiments in order to answer the
following questions: (1) Does using relationship information in
addition to object information actually help the system's performance?
(2) Does the dataset used in this study have exploitable biases,
similar to the findings of Cirik et al.\ \shortcite{Cirik2018b} on the
Google-Ref dataset?  Note that here we use the term ``bias'' to
mean any aspect of the dataset that allows a learning algorithm to
rely on shallow correlations, rather than actually solving the
intended task.  (3) If the dataset does contain biases, how would IRSG
perform on a dataset that did not contain such biases?

\subsection{Comparing IRSG with an Object-Only Baseline} \label{subsec_compare}

To investigate the first two questions, we created a baseline
image-retrieval method that uses information only from object
probabilities.  Given a test image and a scene-graph query, we ran
IRSG's Geodesic Object Proposal method on the test image to obtain
bounding boxes, and we ran IRSG's trained CNN on each bounding box to
obtain a probability for each object category.  For each object
category named in the query, our baseline method simply selects the
bounding box with the highest probability for that query.  No attribute or 
relationship information is used.  We then use a {\it recall at} $k$ ($R@k$)
metric to compare the performance of our baseline method to that of
the IRSG method.

Our $R@k$ metric was calculated as follows.  For a given scene-graph
query, let $S_p$ be the set of {\it positive} images in the test set,
where a positive image is one whose ground-truth object, attribute, and
relationship labels match the query.  Let $S_n$ be the set of negative
images in the test set.  For each scene-graph query, IRSG was run on
both $S_p$ and $S_n$, returning an energy score for each image with
respect to the scene graph.  For each image we also computed a second
score: the geometric mean of the highest object-category
probabilities, as described above.  The latter score ignored attribute and 
relationship information.  We then rank-order each image in the test
set by its score: for the IRSG method, scores (energy values---lower
is better) are ranked in ascending order; for the baseline method,
scores (geometric mean values---higher is better) are ranked in
descending order.  Because the size of $S_p$ is different for
different queries, we consider each positive image $I_p \in S_p$
separately.  We put $I_p$ alone in a pool with all the negative
images, and ask if $I_p$ is ranked in the top $k$.  We define $R@k$ as
the fraction of images in $S_p$ that are top-$k$ in this sense.  For
example, $R@1 = .2$ would mean that 20\% of the positive images are
ranked above all of the negative images for this query; $R@2=.3$ would
mean that 30\% of the positive images are ranked above all but at most one of
the negative images, and so on. This metric is slightly different
from---and, we believe, provides a more useful evaluation than---the
recall at $k$ metric used in \cite{Johnson2015}, which only counted
the position of the top-ranked positive image for each query in
calculating $R@k$.

\begin{figure*}[h]
\center
\includegraphics[width=4.5in]{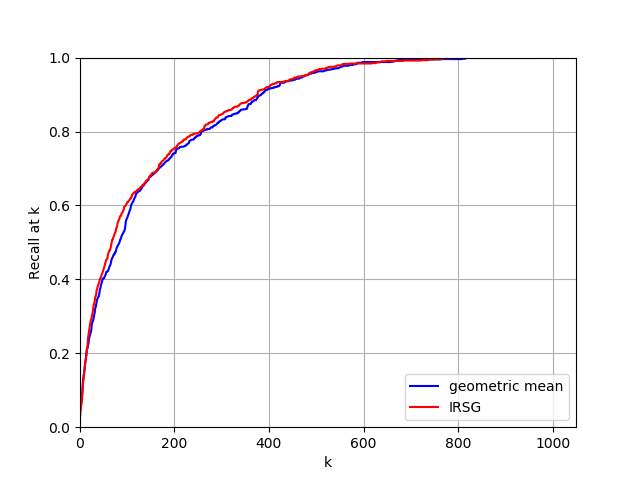}
\caption{Recall at $k$ values for IRSG and the geometric-mean baseline
  on the partial query dataset from \cite{Johnson2015}.  This figure
  shows the averaged $R@k$ values for all partial scene-graph
  queries.} 
\label{original_ratk_compare}
\end{figure*}

We computed $R@k$ in this way for each of the 150 partial scene graphs
that were available in the test set provided by Johnson et al., and
then averaged the 150 values at each $k$.  The results are shown in
Figure~\ref{original_ratk_compare}, for $k=1,...,1000$.  It can be
seen that the two curves are nearly identical.  Our result differs in
a small degree from the results reported in \cite{Johnson2015}, in
which IRSG performed slightly but noticeably better than an
object-only version.  The difference might be due to differences in
the particular subset of scene-graph queries they used (they randomly
selected 119, which were not listed in their paper), or to the
slightly different $R@k$ metrics.

Our results imply that, contrary to expectations, IRSG performance
does not benefit from the system's relationship models.  (IRSG
performance also does not seem to benefit from the system's attribute
models, but here we focus on the role of relationships.)  There are
two possible reasons for this: (1) the object-relationship models
(Gaussian mixture models) in IRSG are not capturing useful
information; or (2) there are biases in the dataset that allow
successful scene-graph grounding without any information from object
relationships.  Our studies show that both hypotheses are correct.

\begin{figure*}[h]
  \center
  \includegraphics[width=4.5in]{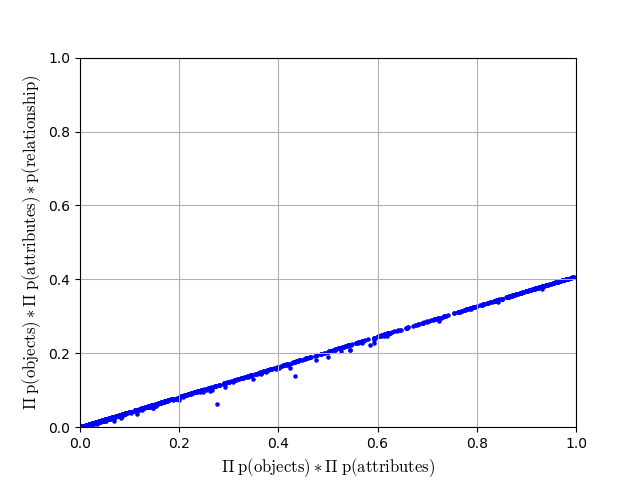}
  \caption{A scatterplot of the factorizations for a single query in
    the original dataset ("clear glasses on woman"), each point
    representing a single image.  The x-axis value is the product of
    the object and attribute probability values from IRSG's
    lowest-energy grounding on this image.  The y-axis value includes
    the product of the relationship probabilities.  A strong
    relationship model would modify the object-attribute factorization and
    create a larger spread of values than what is evident in this
    figure.  We found similar strongly linear relationships for over $90\%$ of the queries in the test set.}
  \label{rel_scatterplot}
\end{figure*}

Figure~\ref{rel_scatterplot} shows results that support the first
hypothesis.  If, for a given scene-graph query, we look at IRSG's
lowest-energy configuration of bounding boxes for every image, and
compare the full (object-attribute-relationship) factorization
(product of probabilities) to the factorization without relationships,
we can see that the amount of information provided by the
relationships is quite small.  For example, for the query ``clear
glasses on woman'', Figure~\ref{rel_scatterplot} is a scatter plot in
which each point represents an image in the test set.  The $x$-axis
values give the products of IRSG-assigned probabilities for objects
and attributes in the scene graph, and the $y$-axis values give the
full product---that is, including the relationship probabilities.  If
the relationship probabilities added useful information, we would
expect a non-linear relationship between the $x$- and $y$-axis values.
However, the plot generally shows a simple linear relationship (linear
regression goodness-of-fit $r^2 = 0.97$), which indicates that the
relationship distribution is not adding significant information to the
final grounding energy.  We found that over $90\%$ of the queries
exhibited very strong linear relationships ($r^2 \geq 0.8$) of this
kind.  This suggests that the relationship probabilities computed by
the GMMs are not capturing useful information.

We investigated the second hypothesis---that there are biases in the
dataset that allow successful object grounding without relationship
information---by a manual inspection of the 150 scene-graph queries
and a sample of the 1,000 test images.  We found two types of such
biases.  In the first type, a positive test image for a given query
contains only one instance of each query object, which makes
relationship information superfluous.  For example, when given a query
such as ``standing man wearing shirt'' there is no need to distinguish
which is the particular ``standing man'' who is wearing a ``shirt'':
there is only one of each.  In the second type of bias, a positive
image for a given query contains multiple instances of the query
objects, but {\it any} of the instances would be a correct grounding
for the query.  For example, when given the query ``black tire on
road'', even if there are many different tires in the image, all of
them are black and all of them are on the road.  Thus any black-tire
grounding will be correct.  Time constraints prevented us from making
a precise count of instances of these biases for each query, but our
sampling suggested that examples of such biases occur in the positive
test images for at least half of the queries.

A closer look at the dataset and queries revealed several additional
issues that make it difficult to evaluate the performance of a visual
grounding system.  While Johnson et al.\ \shortcite{Johnson2015}
reported averages over many partial scene-graph queries, these
averages were biased by the fact that in several cases essentially
same query appeared more than once in the set, sometimes using
synonymous terms (e.g., ``bus on gray street'' and ``bus on gray
road'' are counted as separate queries, as are ``man on bench'' and
``sitting man on bench'').  Removing duplicates of this kind decreases
the original set of 150 queries to 105 unique queries.  Going further,
we found that some queries included two instances of a single object
class: for example, ``standing man next to man''.  We found that when
given such queries, the IRSG system would typically create two
bounding boxes around the same object in the image (e.g., the
``standing man'' and the other man would be grounded as the same
person).

Additionally, there are typically very few positive images per query in the test set.
The mean number of positive images per query is 6.5, and the median
number is 5.  The dataset would benefit from a greater number of
positive results for more thorough testing results.

The dataset was annotated by Amazon Mechanical Turk workers using an
open annotation scheme, rather than directing the workers to select
from a specific set of classes, attributes, and relationships.  Due to
the open scheme, there are numerous errors that affect a system's
learning potential, including mislabeled objects and relationships, as
well as typographical errors (refridgerator [\textit{sic}]), synonyms
(kid/child, man/guy/boy/person), and many prominent objects left
unlabeled.  These errors can lead to false negatives during testing.


\subsection{Testing IRSG on ``Clean'' Queries and Data}
To assess the performance of IRSG without the complications of
many of these data and query issues, we created seven
queries---involving only objects and relationships, no
attributes---that avoided many of the ambiguities described above.  We
made sure that there were at least 10 positive test-set examples for
each query, and we fixed the labeling in the training and test data to
make sure that all objects named in these queries were correctly
labeled.  The queries (and number of positive examples for each in the
test set) are the following:
\begin{itemize}
\item Person Has Beard: 96
\item Person Wearing Helmet: 81
\item Person Wearing Sunglasses: 79
\item Pillow On Couch: 38
\item Person On Skateboard: 29
\item Person On Bench: 18
\item Person On Horse: 13
\end{itemize}
We call this set of queries, along with their training and test examples, the ``clean dataset''. 

\begin{figure*}[h]
  \centering
  \includegraphics[width=4.5in]{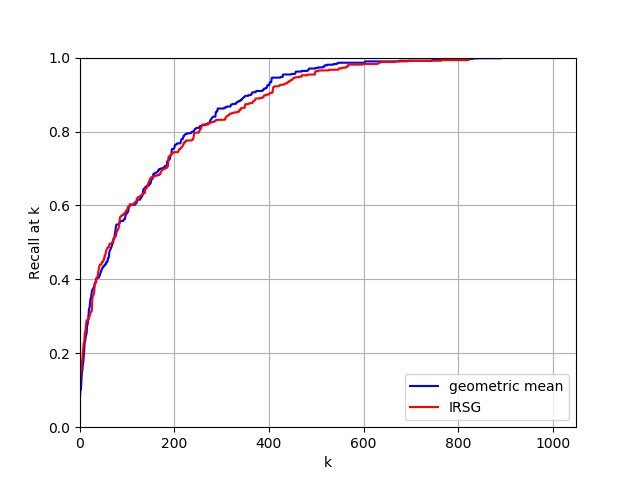}
  \caption{$R@k$ values for the IRSG model and geometric mean model on
    the clean dataset.  This figure shows, for each $k$, the averaged
    $R@k$ values over the seven queries.}
  \label{clean_ratk_compare}
\end{figure*}

Using only these queries, we repeated the comparison between IRSG and
the geometric-mean baseline described above. The $R@k$ results are
shown in Figure~\ref{clean_ratk_compare}.  These results are very
similar to those in Figure~\ref{original_ratk_compare}.  This result
indicates that, while the original dataset exhibits biases and other
problems that make the original system hard to evaluate, it still
seems that relationship probabilities do not provide strongly
distinguishing information to the other components of the IRSG method.
The lack of strong relationship performance was also seen in
\cite{Quinn2018} where the IRSG and object-only baseline method showed
almost identical $R@k$ performance on a different, larger dataset.

\section{Revisiting ``Referring Relationship'' Grounding} 
The IRSG task is closely related to the ``Referring Relationships''
(RR) task, proposed by Krishna et al.\ \shortcite{Krishna2018} and
illustrated in Figure~\ref{RRTask}. The method developed by Krishna et
al.\ uses iterative attention to shift between image regions according
to the given predicate, in order to locate subject and object.  The
authors evaluated their model on several datasets, including the same
images as were in the IRSG dataset (here called ``VRD'' or ``visual
relationship dataset''), but with 4710 referring-relationship queries
(several per test image).  The evaluation metric they reported was
mean {\it intersection over union} (IOU) of the subject and object
detections with ground-truth boxes.  This metric does not give
information about the detection rate.  To investigate whether biases
appear in this dataset and queries similar to the ones we described
above, we again created a baseline method that used only object
information.  In particular, we used the VRD training set to fine-tune
a pre-trained version\footnote{We used
  \texttt{faster\_rcnn\_resnet101\_coco} from
  \url{https://github.com/tensorflow/models/blob/master/research/object_detection/g3doc/detection_model_zoo.md}.}
of the faster-RCNN object-detection method \cite{Ren2015} on the
object categories that appear in the VRD dataset.  We then ran
faster-RCNN on each test image, and for each query selected the
highest-confidence bounding box for the subject and object categories.
(If the query subject and object were the same category, we randomly
assigned subject and object to the highest and second-highest
confidence boxes.)  Finally, for each query, we manually examined
visualizations of the predicted subject and object boxes in each test
image to determine whether the subject and object boxes fit the
subject, object, and predicate of the query.  We found that for 56\%
of the image/query pairs, faster-RCNN had identified correct subject
and object boxes.  In short, our object-only baseline was able to
correctly locate the subject and object 56\% of the time, using no
relationship information.  This indicates significant biases in the
dataset, which calls into question any published
referring-relationship results on this dataset that does not compare
with this baseline.  In future work we plan to replicate the results
reported by Krishna et al.\ \shortcite{Krishna2018} and to compare it
with our object-only baseline.  We hope to do the same for other
published results on referring relationships using the VRD dataset,
among other datasets \cite{Cirik2018,Liu2019,Raboh2019}.

\section{Related Work}
Other groups have explored grounding single objects referred to by
natural-language expressions
\cite{Hu2016,Nagaraja2016,Hu2017,Zhang2018} and grounding all nouns
mentioned in a natural language phrase
\cite{Rohrbach2016,Plummer2017,Plummer2018,Yeh2017}.

Visual grounding is different from, though related to, tasks such as
visual relationship detection \cite{Lu2016}, in which the task is not
to ground a particular phrase in an image, but to detect {\it all}
known relationships.  The VRD dataset we described above is commonly
used in visual relationship detection tasks, and to our knowledge there are no
prior studies of bias and other problems in this dataset.

It should be noted that visual grounding also differs from automated
caption generation \cite{Xu2015} and automated scene graph
generation \cite{Xu2017}, which input an image and output a
natural language phrase or a scene graph, respectively.

The diversity of datasets used in these various studies as well as the
known biases and other problems in many widely used datasets makes it
difficult to determine the state of the art in visual grounding tasks
as well as related tasks such as visual relationship detection.

\section{Conclusions and Future Work}

We have closely investigated one highly cited approach to visual
grounding, the IRSG method of \cite{Johnson2015}.  We demonstrated
that this method does not perform better than a simple object-only
baseline, and does not seem to use information from relationships
between objects, contrary to the authors' claims, at least on the
original dataset of partial scene graphs as well as on our ``clean''
version.  We have also identified exploitable biases and other
problems associated with this dataset, as well as with the version used in
Krishna et al.\ \shortcite{Krishna2018}.

Our work can be seen as a contribution to the effort promoted by Cirik
et al.\ \shortcite{Cirik2018b}: ``to make meaningful progress on grounded
language tasks, we need to pay careful attention to what and how our
models are learning, and whether or datasets contain exploitable
bias.''  In future work, we plan to investigate other prominent
algorithms and datasets for visual grounding, as well as to curate
benchmarks without the biases and problems we described above.  Some
researchers have used synthetically generated data, such as the CLEVR
set \cite{Johnson2017}; however to date the high performances of
visual grounding systems on this dataset have not translated to high
performance on real-world datasets (e.g., Krishna et al.\ \shortcite{Krishna2018}).  We also
plan to explore alternative approaches to visual grounding tasks, such
as the ``active'' approach described by Quinn et al.\ \shortcite{Quinn2018}.

\section*{Acknowledgments}

We are grateful to Justin Johnson of Stanford University for sharing
the source code for the IRSG project, and to NVIDIA corporation for
donation of a GPU used for this work.  We also thank the
anonymous reviewers of this paper for several helpful suggestions for
improvement. This material is based upon work supported by the
National Science Foundation under Grant Number IIS-1423651.  Any
opinions, findings, and conclusions or recommendations expressed in
this material are those of the authors and do not necessarily reflect
the views of the National Science Foundation.


\bibliography{SIVL2019}
\bibliographystyle{acl_natbib}
\end{document}